\documentclass[10pt,twocolumn,letterpaper]{article}

\usepackage[pagenumbers]{cvpr} 

\usepackage{graphicx}
\usepackage{amsmath}
\usepackage{amssymb}
\usepackage{booktabs}

\usepackage{multirow}

\usepackage[pagebackref,breaklinks,colorlinks]{hyperref}

\usepackage[capitalize]{cleveref}
\crefname{section}{Sec.}{Secs.}
\Crefname{section}{Section}{Sections}
\Crefname{table}{Table}{Tables}
\crefname{table}{Tab.}{Tabs.}

\def\cvprPaperID{867} 
\def\confName{CVPR}
\def\confYear{2022}

\begin{document}

\title{Adaptive Shrink-Mask for Text Detection}

\author{Chuang Yang,~~Mulin Chen,~~Yuan Yuan,~~Qi Wang*,~~Xuelong Li\\
The School of Computer Science, and the School of Artificial Intelligence,\\
Optics and Electronics (iOPEN),\\
Northwestern Polytechnical University, Xian 710072, Shaanxi, P. R. China.
}
\maketitle

\begin{abstract}
Existing real-time text detectors reconstruct text contours by shrink-masks directly, which simplifies the framework and can make the model run fast. However, the strong dependence on predicted shrink-masks leads to unstable detection results. Moreover, the discrimination of shrink-masks is a pixelwise prediction task. Supervising the network by shrink-masks only will lose much semantic context, which leads to the false detection of shrink-masks. To address these problems, we construct an efficient text detection network, Adaptive Shrink-Mask for Text Detection (ASMTD), which improves the accuracy during training and reduces the complexity of the inference process. At first, the Adaptive Shrink-Mask (ASM) is proposed to represent texts by shrink-masks and independent adaptive offsets. It weakens the coupling of texts to shrink-masks, which improves the robustness of detection results. Then, the Super-pixel Window (SPW) is designed to supervise the network. It utilizes the surroundings of each pixel to improve the reliability of predicted shrink-masks and does not appear during testing. In the end, a lightweight feature merging branch is constructed to reduce the computational cost. As demonstrated in the experiments, our method is superior to existing state-of-the-art (SOTA) methods in both detection accuracy and speed on multiple benchmarks.
\end{abstract}

\section{Introduction}
Text detection, a key technology to provide text location information for many scene text recognition (STR)~\cite{nguyen2021dictionary,bhunia2021towards,litman2020scatter} related applications, has become a hot topic recently. Existing text detection methods can be roughly divided into non-real-time methods and real-time methods. The former~\cite{DBLP:conf/cvpr/FengYZL21,zhu2021fourier,DBLP:conf/cvpr/ZhangZHLYWY20} focuses on achieving high detection accuracy. However, since the complex framework leads to low detection speed and high memory requirement, non-real-time methods are hard to deploy in the device. Though the latter~\cite{wang2019efficient,liao2020real} can run fast with a lightweight network and simple post-processing, the limited detection accuracy restricts their applicability. Therefore, how to design an efficient framework to detect arbitrary-shaped text instances with high detection accuracy and speed is still explored.

\begin{figure}
	\begin{center}
		\includegraphics[width=0.4\textwidth]{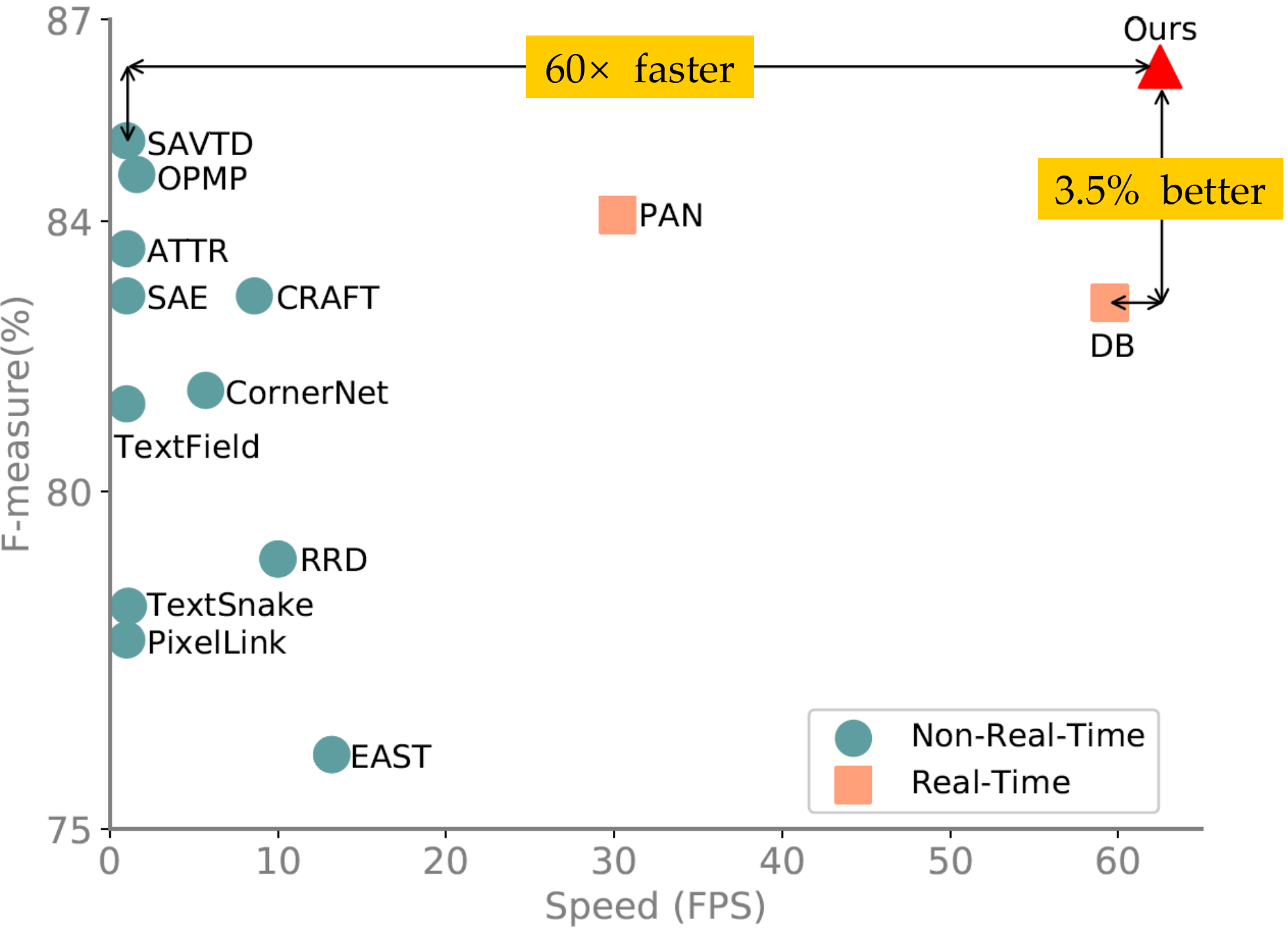}
	\end{center}
	\vspace{-5mm}
	\caption{Performance of detection accuracy and speed on MSRA-TD500 dataset. Our method outperforms the best non-real-time method SAVTD~\cite{DBLP:conf/cvpr/FengYZL21} in detection accuracy and runs faster than the best real-time method DB~\cite{liao2020real}.}
	\label{V1}
	\vspace{-3mm}
\end{figure}

Considering the aforementioned problems, we construct a novel real-time text detection network based on adaptive shrink-mask (ASM) and super-pixel window (SPW), namely ASMTD, which can achieve comparable detection accuracy with non-real-time methods and detection speed with real-time methods. Specifically, the proposed ASM represents text instances by shrink-masks and adaptive offsets. In the inference stage, text contours can be rebuilt by extending the shrink-mask contours outward by adaptive offsets directly, which simplifies the post-processing and makes it work that run faster and deploy easier. At the same time, adaptive offsets can evaluate the distance between shrink-mask contours and text contours accurately even when the predicted shrink-masks deviate from the corresponding ground-truth, which brings significant improvement for the accuracy of rebuilt text contours. Moreover, SPW is proposed to improve the accuracy of pixelwise prediction tasks. It encourages the network to provide rich semantic context for the discrimination of shrink-masks and the prediction of adaptive offsets. Importantly, the SPW can be removed from the inference process, which does not introduce any temporal cost. Additionally, a lightweight feature merging branch is constructed to save computational cost to further speed up the inference process. The experiment results demonstrate our method is superior to existing state-of-the-art (SOTA) text detection methods (as shown in Fig.~\ref{V1}). The contributions of this work are summarized as follows:

\begin{itemize}
	\item[1.] An Adaptive Shrink-Mask is proposed to rebuild text contour by the shrink-mask and adaptive offset, which ensures a simple framework and can reconstruct text contour accurately even when the predicted shrink-mask deviates from the corresponding ground-truth.
	
	\item[2.] Super-pixel Window is proposed to encourage the network to provide rich semantic context for pixelwise prediction tasks, which helps to recognize the shrink-mask and predict the adaptive offset and brings no extra computational cost to the inference process.
	
	\item[3.] An efficient detector with a lightweight network and simple post-processing is proposed. It makes detection results more reliable, run faster, and deployment easier, which provides essential support for applications.
	
\end{itemize}

\section{Related Work}
\label{Related Work}
In recent years, scene text detection technology is rapidly developing. Existing text detection methods can be roughly classified into non-real-time and real-time methods.

\subsection{Non-Real-Time Text Detection Methods.}
Non-real-time text detection methods are composed of complicated network and post-processing and focus on achieving high detection accuracy. Liao~\etal\cite{liao2018rotation} proposed rotation-sensitive features for detecting oriented text instances. Zhou~\etal\cite{zhou2017east} proposed dense prediction for abandoning the anchor mechanism. Law and Deng~\cite{law2018cornernet} modeled text instances by four corner regions and combined them to rebuild text regions. However, they failed to detect irregular-shaped text instances. Though Wang~\etal\cite{wang2020textray} represented text contours by multiple contour points, highly curved text contours could not be fitted accurately. Some works~\cite{ma2021relatext,long2018textsnake,baek2019character} decomposed text instances into a number of character-level boxes and connected them to rebuild text contours. Zhu~\etal\cite{zhu2021fourier} converted text instance contours from point sequences into Fourier signature vectors. Liu~\etal\cite{liu2020abcnet} represented text contours by Bezier-Curve. Zhang~\etal\cite{zhang2020opmp} detected text instances by an effective pyramid lengthwise and sidewise residual sequence model. Zhang~\etal\cite{zhang2019look} and Wang~\etal\cite{wang2020contournet} predicted quadrilateral bounding boxes at first and then segmented text regions in the range of the boxes. To improve the detection performance of very long sentences, Tian~\etal\cite{tian2019learning} segmented an embedding map to connect multiple tiny text instances as integral ones. Xu~\etal\cite{xu2019textfield} predicted pixel classes and directions to detect text from the background and distinguish different text instances, respectively. Although these methods can fit arbitrary-shaped text contours accurately, the low detection speed and high memory requirement hinder providing support for STR-related~\cite{karthick2019steps,baek2019wrong,ingle2019scalable} applications.

\begin{figure*}
	\begin{center}
		\includegraphics[width=0.97\textwidth]{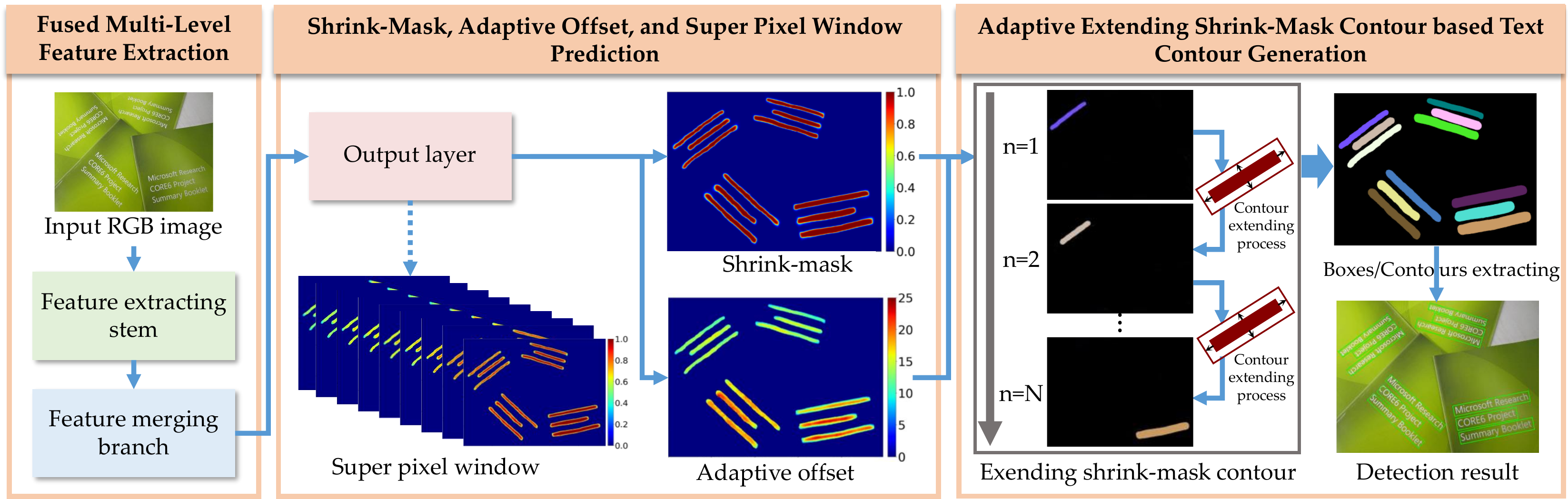}
	\end{center}
	\vspace{-5mm}
	\caption{Overall architecture of ASMTD. Dashed arrow is the training only operator. ``n=1, n=2, ..., n=N'' indicates the $1$th, $2$th, ..., $N$th shrink-mask in the image, respectively.}
	\label{V2}
\end{figure*}

\subsection{Real-Time Text Detection Methods.}
To obtain fast detection speed and reduce memory requirements, many real-time text detection methods are proposed recently. These methods are equipped with a lightweight backbone and adopt the shrink-mask based segmentation framework to detect text instances. For example, Wang~\etal\cite{wang2019efficient} and Liao~\etal\cite{liao2020real} rebuilt text contours by extending shrink-mask regions. Specifically, Wang~\etal\cite{wang2019efficient} segmented shrink-masks and text masks simultaneously. In the inference stage, the authors extended shrink-masks to text masks through pixel-level post-processing, which was time-consuming. Since Liao~\etal\cite{liao2020real} could compute the distances between shrink-masks and text masks according to shrink-masks through the algorithm proposed in \cite{vatti1992generic}, this method only needed to segment shrink-masks and extend them through patch-level post-processing, which further improved the detection speed. However, though these existing real-time text detection methods enjoy fast detection speed, the detection accuracy is still far behind non-real-time methods.

\section{Methodology}
In this section, the overall architecture of ASMTD is introduced firstly. Then, the adaptive shrink-mask (ASM) and super-pixel window (SPW) are described in detail, respectively. Next, the label generation process is illustrated. In the end, the optimization function is elaborated.

\begin{figure}
	\begin{center}
		\includegraphics[width=0.43\textwidth]{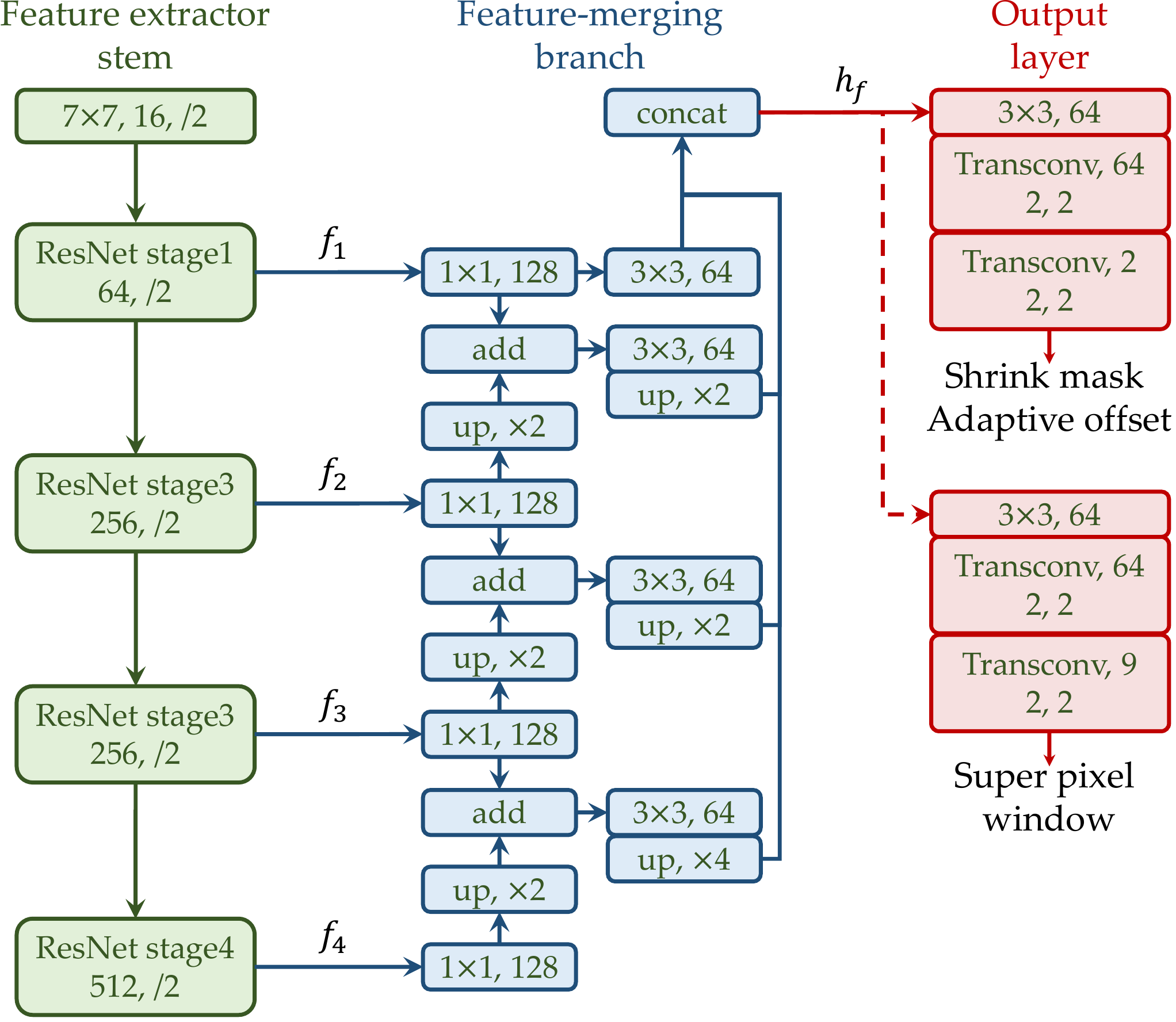}
	\end{center}
	\vspace{-5mm}
	\caption{Details of feature extracting network. Red dashed arrow is the training only operator.}
	\label{V3}
	\vspace{-3mm}
\end{figure}
\subsection{Overall Architecture}
\label{architecture}
The overall architecture of ASMTD is shown in Fig.~\ref{V2}, which consists of feature extracting stem, feature merging branch, output layer, and adaptive extending strategy based post-processing. In the Inference stage, a fused multi-level feature map is extracted through feature extracting stem and feature merging branch at first. Then, the output layer conducts on the fused feature map to predict the shrink-mask, adaptive offset, and SPW. The SPW is a training only operator, which helps to provide rich context information for the prediction tasks and brings no extra computational cost to the inference process. For shrink-mask and adaptive offset, the pixel value denotes the probability of whether it belongs to the text instance and the minimum distance between the pixel and text contour, respectively. In the end, to reconstruct all text contours in the image, all shrink-mask contours are extended by the predicted adaptive offsets through post-processing one by one. Benefiting from the advantages of adaptive offset, our method can rebuild text contours even when the predicted shrink-mask deviates from the ground-truth, which brings significant improvement for detection accuracy.

The details of feature extracting stem, feature merging branch, and output layer are illustrated in Fig.~\ref{V3}. ResNet~\cite{he2016deep} is adopted as feature extracting stem directly. It generates multi-level feature maps whose sizes are $\frac{1}{4}$,$\frac{1}{8}$,$\frac{1}{16}$ and $\frac{1}{32}$ of the input image, respectively. To reduce the computational cost to speed up the inference process, we follow the idea of Feature Pyramid Network (FPN)~\cite{lin2017feature} and construct a lightweight feature merging branch, which is used for concatenating the multi-level feature maps to generate a fused feature map. The output layer is responsible for predicting the shrink-mask, adaptive offset, and SPW.

\begin{figure}
	\begin{center}
		\includegraphics[width=0.45\textwidth]{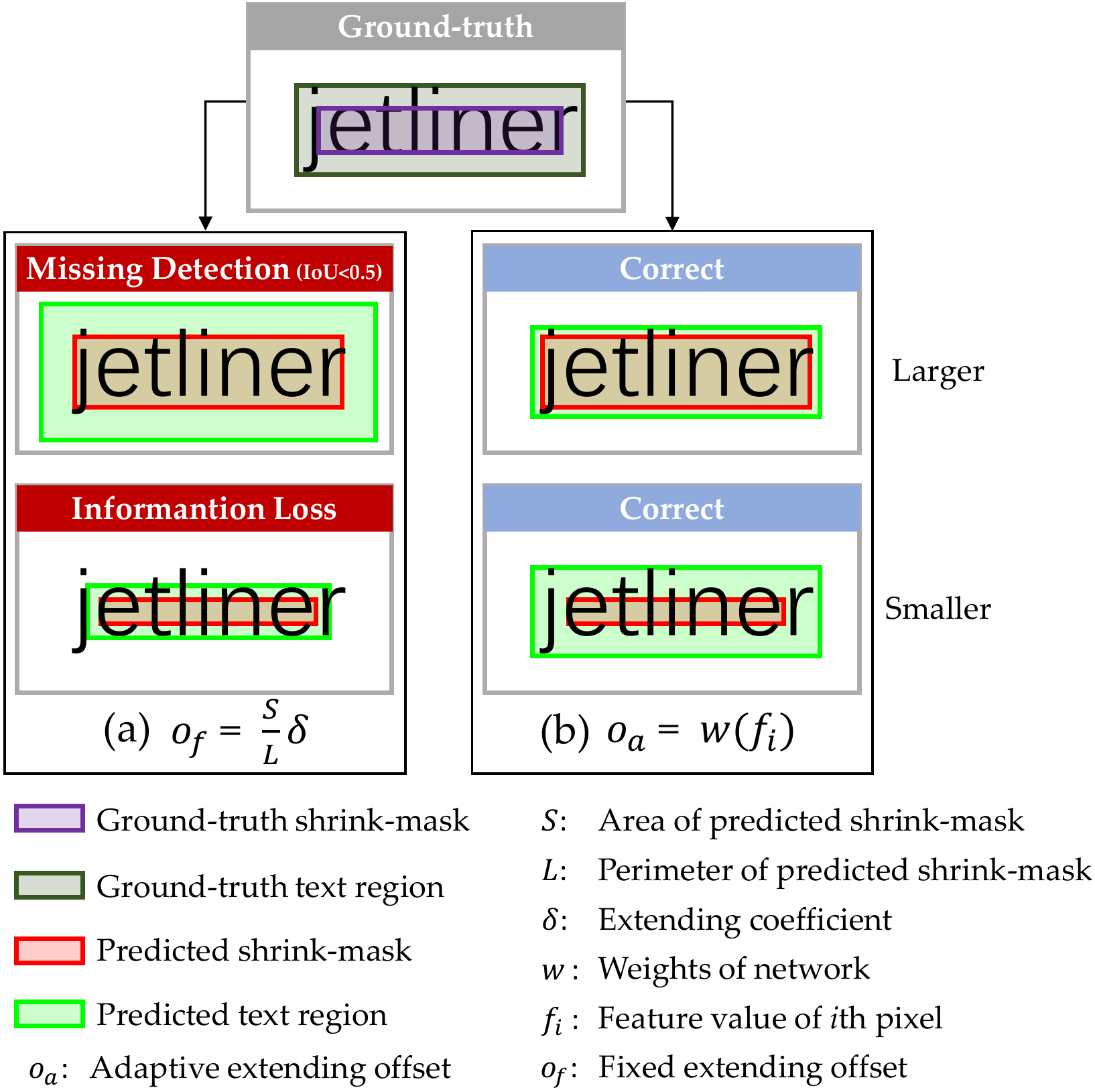}
	\end{center}
	\vspace{-5mm}
	\caption{Illustration of essential differences between traditional fixed extending strategy and adaptive extending strategy.}
	\label{V4}
	\vspace{-3mm}
\end{figure}

\subsection{Adaptive Shrink-Mask}
\label{asm}
To avoid missing detection and improve the information integrity of detected text instances, we propose an adaptive shrink-mask (ASM) to represent text instances. ASM based model can rebuild text contours by extending shrink-mask contours outward by adaptive offsets accurately, even when the predicted shrink-mask deviates from the ground-truth.

As we can see from Fig.~\ref{V4}~(a), existing real-time methods (such as~\cite{liao2020real}) rebuild text contour through extending the predicted shrink-mask contour outward by a fixed offset $o_{f}$: 
\begin{eqnarray}
\label{E1}
o_{f}=\frac{S_s}{L_s}\delta_t,
\end{eqnarray}
where $S_s$ and $L_s$ are the area and perimeter of the predicted shrink-mask, respectively. $\delta_t$ is the extending coefficient. 

As shown in Eq.~(\ref{E1}), Since $o_{f}$ deeply depends on the area and perimeter of the predicted shrink-mask, the $o_{f}$ will deviate from the ground-truth a large when the predicted shrink-mask is larger or smaller than the ground-truth, which further leads to missing detection or text information loss (Fig.~\ref{V4}~(a)) and influences model detection accuracy. Considering these problems, we propose to extend the shrink-mask contour by an adaptive offset $o_a$ :
\begin{eqnarray}
\label{E2}
o_a=w\left( f_i \right),
\end{eqnarray}
where $w(\cdot)$ denotes the weights of the output layer in Fig.~\ref{V3}. $f_i$ means the $i$th pixel value of the fused feature map ($h_f$ in Fig.~\ref{V3}). Different from $o_{f}$, $o_{a}$ is independent of the predicted shrink-mask, which means the distance between predicted shrink-mask contour and text contour can be evaluated accurately, even when the predicted shrink-mask deviates from the corresponding ground-truth. Benefiting from the advantages of $o_{a}$, our method can avoid missing detection and improve text information integrity when rebuilding text contour based on incorrect shrink-mask (Fig.~\ref{V4}~(b)), which brings significant improvement for detection accuracy. Moreover, since the text contours can be reconstructed by extending the shrink-mask contour outward directly, the ASMTD enjoys a simpler network and post-processing than non-real-time methods (such as~\cite{zhu2021fourier,DBLP:conf/cvpr/ZhangZHLYWY20}), which facilitates our model run times faster.

\begin{figure}
	\begin{center}
		\includegraphics[width=0.45\textwidth]{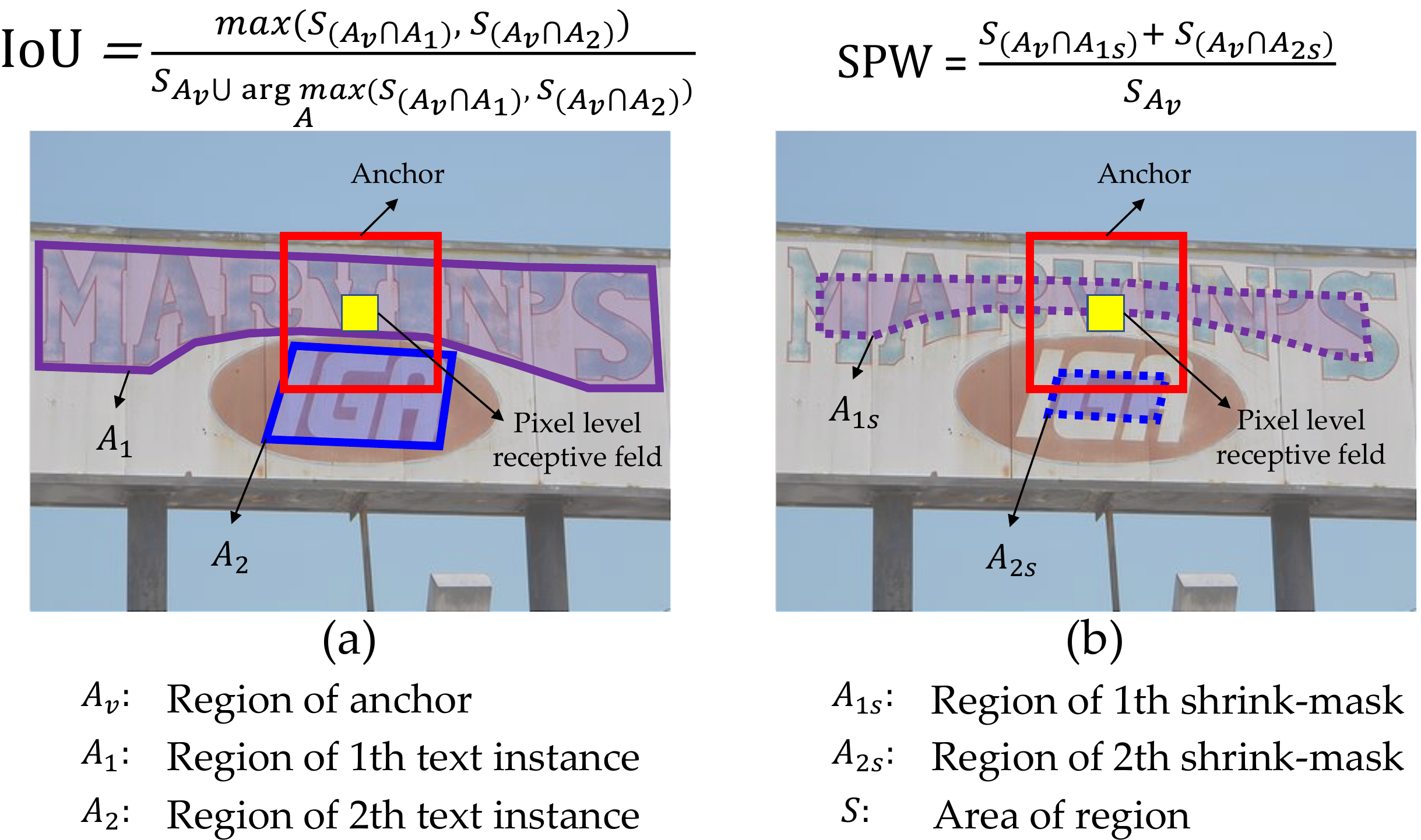}
	\end{center}
	\vspace{-5mm}
	\caption{Illustration of essential differences between Intersection of Union (IoU) and Super-pixel Window (SPW).}
	\label{V5}
	\vspace{-3mm}
\end{figure}

\subsection{Super-pixel Window}
Since some background regions enjoy similar low-level features (such as color, texture, and gradients) with text instances, existing real-time methods are hard to distinguish the shrink-mask from them according to the pixel-level features (the yellow regions in Fig.~\ref{V5}). To provide rich semantic context for the pixelwise prediction tasks, we propose the Super-pixel Window (SPW) to train our model. 

The existing way to obtain rich semantic context is to optimize the network under the supervision of the intersection of union (IoU). As shown in Fig.~\ref{V5}, the IoU is defined as:
\begin{eqnarray}
\label{E3}
{\rm IoU}=\frac{max\left( S_{\left( A_v\cap A_1 \right)},S_{\left( A_v\cap A_2 \right)} \right)}{S_{A_v\cup \underset{A}{argmax}\left( S_{\left( A_v\cap A_1 \right)},S_{\left( A_v\cap A_2 \right)} \right)}},
\end{eqnarray}
where $A_v$ indicates the anchor region. $A_1$ and $A_2$ are the regions of text instances. $\cap$ and $\cup$ are the intersection and union operators. $S_{(\cdot)}$ is the area of region. $\underset{A}{argmax}(\cdot)$ is the text region that enjoys maximum intersection with $A_v$. 

However, since IoU considers the whole text region, it brings interference for the training process when meeting some very long text instances (such as $A_1$ in Fig.~\ref{V5}~(a)) that are far beyond the network receptive field. Moreover, smaller text instances are treated as background for IoU, which further leads to semantic ambiguity between text and background. Considering these problems, we propose SPW:
\begin{eqnarray}
\label{E4}
{\rm SPW}=\frac{S_{\left( A_v\cap A_{1s} \right)}+S_{\left( A_v\cap A_{2s} \right)}}{S_{A_v}},
\end{eqnarray}
where $A_{1s}$ and $A_{2s}$ are the regions of shrink-masks. 

Different from IoU, as shown in Fig.~\ref{V5}~(b), the proposed SPW enjoys the following advantages: (1) only the shrink-masks within the range of anchor are recognized, which effectively avoids the interference brought by very long text instances. At the same time, SPW treats the interval region between shrink-mask contour and text contour as background, which helps to distinguish shrink-masks and predict adaptive offsets; (2) all shrink-masks within the range of anchor are considered no matter their scales are large or small, which avoids the semantic ambiguity brought by treating the small ones as background. Additionally, the SPW brings no extra computational cost because it can be removed from the inference process, which ensures high detection speed. 

\begin{figure}
	\begin{center}
		\includegraphics[width=0.45\textwidth]{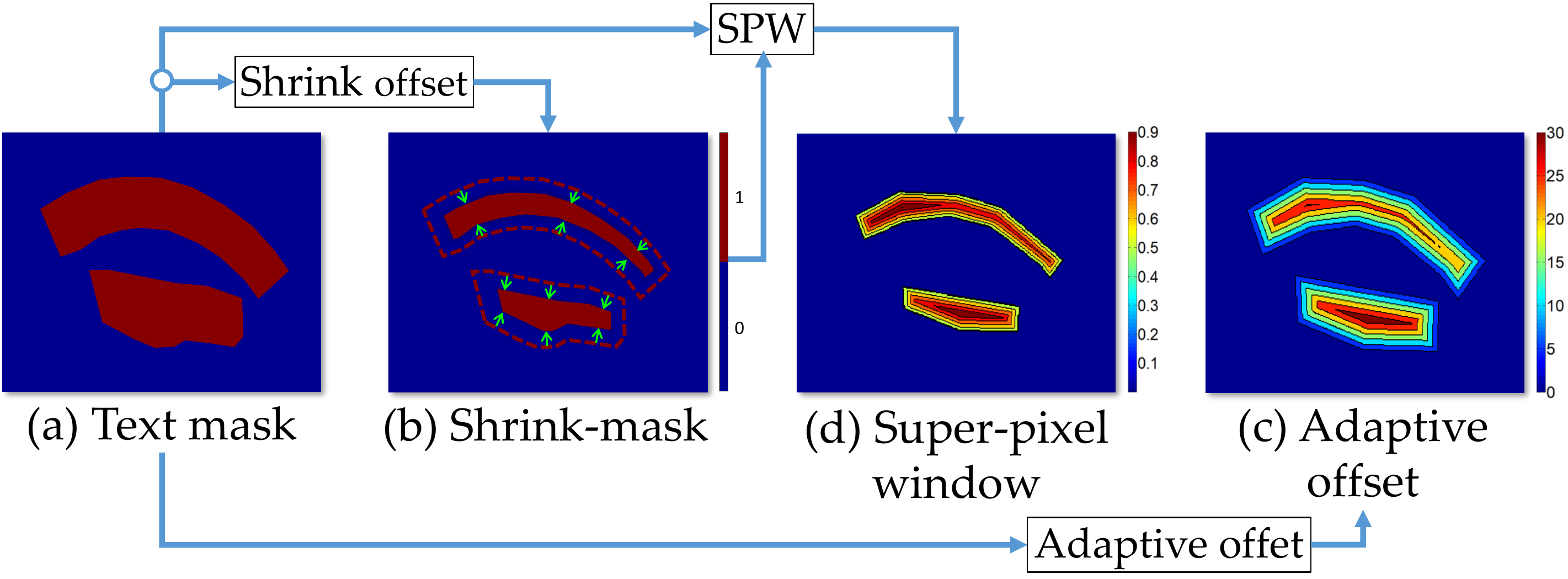}
	\end{center}
	\vspace{-5mm}
	\caption{Illustration of the label generation process.}
	\label{V6}
	\vspace{-3mm}
\end{figure}

\subsection{Label Generation}
The label generation processes for shrink-mask map, adaptive offset map, and SPW map are visualized in Fig.~\ref{V6}.

For shrink-mask (as shown in Fig.~\ref{V6}~(b)), the corresponding contour is generated through moving the text instance contour inward by shrink offset $o_s$ that is computed by the Vatti clipping algorithm~\cite{vatti1992generic}:
\begin{eqnarray}
\label{E5}
o_s=\frac{S_t}{L_t}\left( 1-\delta_s ^2 \right),
\end{eqnarray}
where $S_t$ and $L_t$ are the area and perimeter of text instance, respectively. $\delta_s$ is the shrinking coefficient, which is set to 0.4 empirically.

For adaptive offset $o_a$ (as shown in Fig.~\ref{V6}~(c)), which is defined as the minimum distance between the pixel in the range of text instance and text contour:
\begin{eqnarray}
\label{E6}
o_a=min \left\{ \left\| p-p_m \right\| _{2}^{2} \right\},~m=1,2,...,M,
\end{eqnarray}
where $p$ and $p_m$ are the coordinates of pixels of text instance and text contour, respectively. $min(\cdot)$ indicates the minimum operator. $M$ is the number of contour points.

For SPW (as shown in Fig.~\ref{V6}~(d)), it treats shrink-masks as valid regions and the corresponding pixel values of the map are computed by the Eq.~(\ref{E4}).

\subsection{Optimization}
The proposed ASMTD can be regarded as a multi-task network aiming at shrink-mask segmentation, adaptive offset prediction, and SPW prediction. For shrink-mask segmentation, the classic dice loss~\cite{DBLP:conf/3dim/MilletariNA16} is used for evaluating the difference between the segmentation binary masks and the corresponding labels, which is formulated as:
\begin{eqnarray}
L_{sm}=1-\frac{2\times |Y\cap \widehat{Y}|+1}{|Y|+|\widehat{Y}|+1},
\end{eqnarray}  
where $Y$ and $\widehat{Y}$ are the ground-truth and predicted shrink-masks. Since the serious imbalance of positive (text instances) and negative samples (background) and plenty of simple samples, Online Hard Example Mining (OHEM)~\cite{shrivastava2016training} is adopted when calculating $L_{sm}$.

For adaptive offset and SPW prediction, the ratio loss $L_{ratio}$ is adopted to compute their gradients:
\begin{eqnarray}
L_{ratio}(P,~\widehat{P})=\log \frac{max \left( P,~\widehat{P} \right)}{min \left( P,~\widehat{P} \right)},
\end{eqnarray}       
where $P$ and $\widehat{P}$ are the ground-truth and prediction, respectively. Therefore, $L_{ratio}$ based adaptive offset loss $L_{o_a}$ and SPW loss $L_{SPW}$ can be defined as:
\begin{eqnarray}
L_{o_a}=L_{ratio}(o_a,~\widehat{o_a}),
\end{eqnarray}  
\begin{eqnarray}
L_{SPW}=L_{ratio}(SPW,~\widehat{SPW}),
\end{eqnarray}  
where $o_a$ and $\widehat{o_a}$ are the ground-truth (computed by the Eq.~(\ref{E6})) and predicted $o_a$. $SPW$ and $\widehat{SPW}$ are the ground-truth (computed by the Eq.~(\ref{E4})) and predicted $SPW$. 

The final loss function $L$ used for training the proposed network is given by:
\begin{eqnarray}
L=\lambda _1L_{sm}+\lambda _2L_{o_a}+\lambda _3L_{SPW},
\end{eqnarray}  
where $\lambda$ is hyper-parameter and used to balance multiple losses weights. In this paper, $\lambda _1$, $\lambda _2$, and $\lambda _3$ are set to 1, 0.25, and 0.25 respectively.

\section{Experiments}
\label{Experiments}
To demonstrate the effectiveness of the ASM and SPW, we conduct ablation studies on the MSRA-TD500 and Total-Text datasets. Moreover, we also compare the proposed ASMTD with related state-of-the-art (SOTA) methods on multiple public datasets to show the superiority in both detection accuracy and speed.

\subsection{Datasets}
\textbf{SynthText}~\cite{gupta2016synthetic} is composed of 800k synthetic images generated by combining varied text instances with 8k natural images. This dataset is used for pre-training the proposed ASMTD. \textbf{MSRA-TD500} \cite{yao2012detecting} consists of arbitrary-oriented and long text sentences. It includes 300 training images and 200 test images. Since the training images are rather less, we include 400 images from HUST-TR400 \cite{yao2014unified} as training data. \textbf{Total-Text} \cite{ch2017total} contains various shapes word-level text instances (such as horizontal, multi-oriented, and curved shapes) simultaneously. It is composed of 1255 training images and 300 testing images. \textbf{CTW1500} \cite{yuliang2017detecting} is a dataset for testing the model performance on line-level curved text instances, which has 1000 training images and 500 testing images.

\subsection{Implementation Details}
The feature extracting stem is pre-trained on ImageNet \cite{deng2009imagenet} and the whole network is pre-trined on SynthText~\cite{gupta2016synthetic}. In the fine-tuning stage, Adam~\cite{kingma2014adam} is employed to train the model. The initial learning rate is set to 0.001 and is adjusted by the `poly' strategy used in~\cite{zhao2017pyramid}. The training batch size is set to 16. We initialize the weights of feature merging branch and output layers with the strategy in \cite{he2015delving}. For all datasets, the blurred text regions labeled as DO NOT CARE are ignored during the training. All the experiments are conducted on Pytorch using a workstation with two 1080Ti GPUs. To increase the training data and avoid over-fitting, we adopt the following data augmentation strategies: (1) random horizontal flipping, (2) random scaling and cropping, (3) random rotation with an angle range of (-10°, 10°).

\begin{table*}[t]
	\centering
	\begin{tabular}{c|c|cc|ccc|c}
		\hline
		\multicolumn{1}{l|}{} & \multicolumn{7}{c}{Image scale for testing~ (S~:~736)}                 \\ \hline
		\#                     &           & ASM & SPW  & Precision (\%) & Recall (\%) & F-measure (\%) & FPS  \\ \hline
		1                      & baseline  &     &     & 87.2      & 81.6   & 84.3      & 64.2 \\ \hline
		2                      & baseline+ & \checkmark   &     & 88.9      & 82.5   & 85.6    & 62.5 \\ \hline
		3                      & baseline+ & \checkmark   & \checkmark     & 89.8      & 83.1   & 86.3    & 62.5 \\ \hline
	
	\end{tabular}
	\vspace{-3mm}
	\caption{Detection results with different settings on MSRA-TD500. ``S: 736'' means that the shorter side of each testing image is resized to be 736 pixels. ``baseline'' means the framework equipped with traditional shrink-mask only. ``ASM'' indiates adaptive shrink-mask. ``SPW'' means super-pixel window.}
	\label{ablation1}
	\vspace{-3mm}
\end{table*}

\subsection{Ablation Study}
The ablation study is conducted to show the effectiveness of the proposed Adaptive Shrink-Mask (ASM) and Super-pixel Window (SPW).

\begin{figure}
	\begin{center}
		\includegraphics[width=0.45\textwidth]{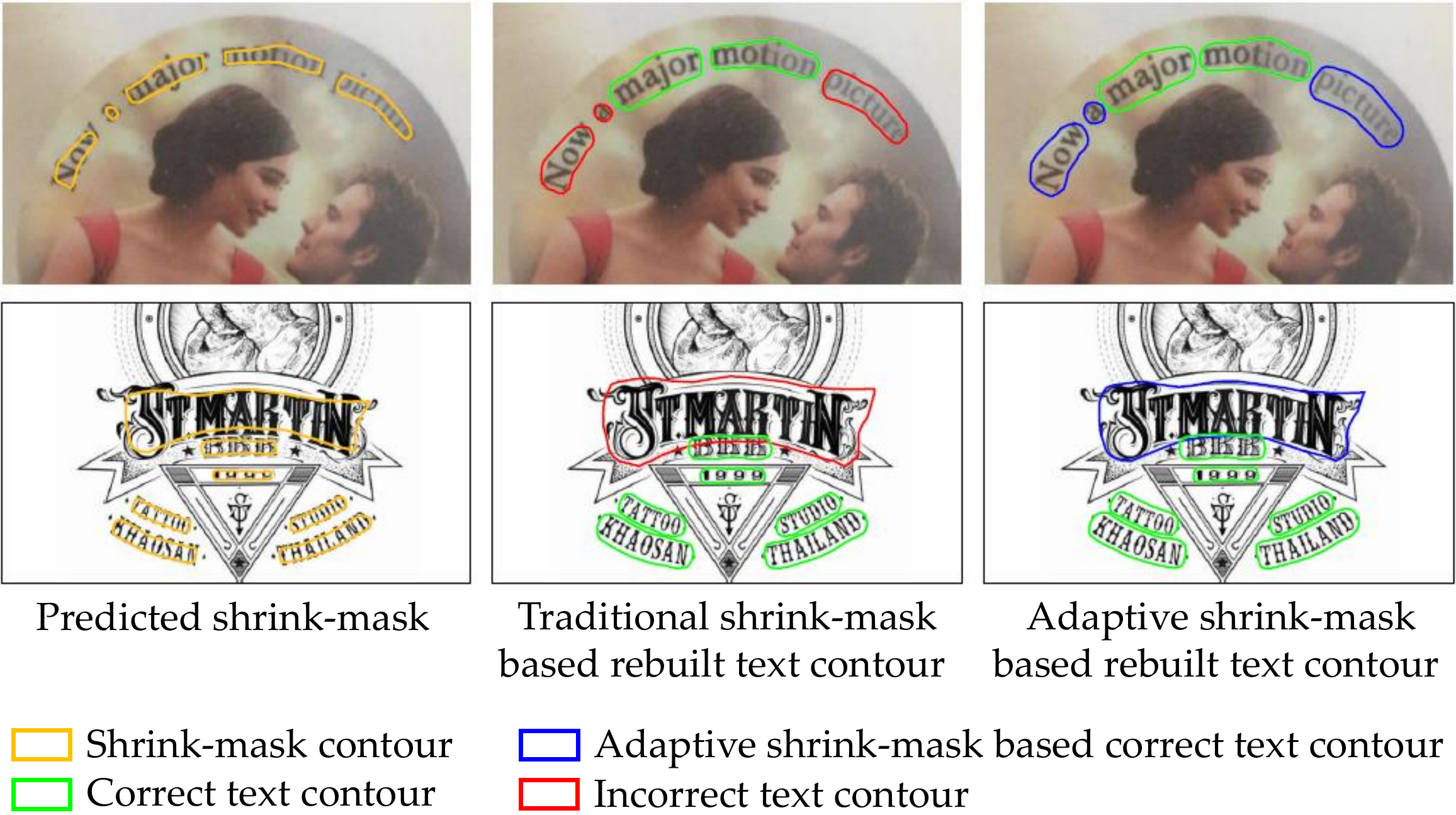}
	\end{center}
	\vspace{-5mm}
	\caption{Qualitative comparisons with traditional shrink-masks based rebuilt text contours and adaptive shrink-masks based rebuilt text contours.}
	\label{V7}
	\vspace{-3mm}
\end{figure}

\textbf{Adaptive Shrink-Mask.} As shown in Tab.~\ref{ablation1}~\#2, since the proposed ASM can avoid missing detection effectively, which brings 1.3\% improvement in terms of F-measure on MSRA-TD500. At the same time, we evaluate our method on Total-Text benchmark with different settings. Specifically, in Tab.~\ref{ablation2}, ASM achieves 1.2\% and 1.1\% gain in F-measure when IoU are set as 50\% and 75\% respectively. Particularly, for the ASM based model, the corresponding F-measure with 75\% IoU achieves 85.5\%, which outperforms the F-measure of baseline with 50\% IoU by 0.6\%, which shows the ASM can improve the integrity of rebuilt text contours. Moreover, we show some examples in Fig.~\ref{V7}. Traditional shrink-masks based rebuilt text contours (Fig.~\ref{V7}~red colored geometries) are smaller than ground-truth, which leads to text information loss. Adaptive shrink-masks based rebuilt text contours (Fig.~\ref{V7}~blue colored geometries) avoid these problems effectively. In Fig.~\ref{V7}~second~row, adaptive shrink-masks based rebuilt text contours are accurate, even when the shrink-masks are larger than ground-truth. These experiments demonstrate the superiority of the proposed ASM for detecting text instances.

\begin{table}
	\setlength{\tabcolsep}{0.25mm}
	\begin{tabular}{c|ccc|ccc}
		\hline
		\multirow{2}{*}{} & \multicolumn{6}{c}{TotalText}  \\ \cline{2-7}
		& $\rm {P}_{50}$(\%) & $\rm {R}_{50}$(\%) & $\rm {F}_{50}$(\%) & $\rm {P}_{75}$(\%) & $\rm {R}_{75}$(\%) & $\rm {F}_{75}$(\%) \\ \hline
		w/o ASM  &  87.6  & 82.3  & 84.9  & 86.6  & 82.3  &  84.4 \\ \hline
		with ASM & 88.5  & 83.8  &  86.1 & 87.3  & 83.8  &  85.5  \\ \hline
	\end{tabular}
	\vspace{-3mm}
	\caption{Detection results with different settings on Total-Text. ``ASM'' indiates adaptive shrink-mask. ``$\rm P_{50}$'', ``$\rm R_{50}$'', and ``$\rm F_{50}$'' indicate the precision, recall, and f-measure that IoU is set as 50\%. ``$\rm P_{75}$'', ``$\rm R_{75}$'', and ``$\rm F_{75}$'' are the precision, recall, and f-measure that IoU is set as 75\%.}
	\centering
	\label{ablation2}
\end{table}

\textbf{Super-pixel Window.} Benefiting from the rich semantic context brought by SPW, our method achieves significant performance gain in terms of F-measure. As shown in Tab.~\ref{ablation1}~\#3, the ASM and SPW bring 2\% improvement in F-measure. Moreover, since the SPW does not participate in the reconstructing process of text contours, the model equipped with SPW brings no extra computational cost to the inference process (as shown in Tab.~\ref{ablation1}~\#2--\#3). The experiment results prove the SPW not only can improve the detection accuracy but also ensures high detection speed.

\begin{table}
	\centering
	\setlength{\tabcolsep}{1.7mm}
	\begin{tabular}{c|c|cccc}
		\hline
		Type & Methods & P (\%) & R (\%) & F (\%)& FPS \\ \hline
		& EAST~\cite{zhou2017east} & 87.3 & 67.4 & 76.1 & \textcolor{blue}{\textbf{13.2}} \\ \cline{2-6}
		& PixelLink~\cite{DBLP:conf/aaai/DengLLC18} & 83.0 & 73.2 & 77.8 & - \\ \cline{2-6} 
		& TextSnake~\cite{long2018textsnake} & 83.2 & 73.9 & 78.3 & 1.1 \\ \cline{2-6} 
		& RRD~\cite{liao2018rotation} & 87.0 & 73.0 & 79.0 & 10 \\ \cline{2-6}
		\multirow{2}{*}{NRT}
		& TextField~\cite{xu2019textfield} & 87.4 & 75.9 & 81.3 & - \\ \cline{2-6} 
		& CornerNet~\cite{law2018cornernet} & 87.6 & 76.2 & 81.5 & 5.7 \\ \cline{2-6} 
		& CRAFT~\cite{baek2019character} & 88.2 & 78.2  & 82.9 & 8.6 \\  \cline{2-6} 
		& SAE~\cite{tian2019learning} & 84.2 & 81.7 & 82.9 & - \\  \cline{2-6} 
		& OPMP~\cite{zhang2020opmp} & 86.0 & 83.4 & 84.7 & 1.6 \\ \cline{2-6} 
		& SAVTD~\cite{DBLP:conf/cvpr/FengYZL21} & 89.2 & 81.5 & \textcolor{blue}{\textbf{85.2}} & - \\ \hline
		& DB~\cite{liao2020real}& 90.4 & 76.3 & 82.8 & 59.5*  \\ \cline{2-6} 
		RT
		& PAN~\cite{wang2019efficient} & 84.4 & 83.8 & 84.1 & 30.2 \\ \cline{2-6}
		& \textbf{Ours} & 89.8 & 83.1 & \textcolor{red}{\textbf{86.3}} & \textcolor{red}{\textbf{62.5}} \\ \hline
		
	\end{tabular}
	\vspace{-3mm}
	\caption{Comparison with related methods on MSRA-TD500. ``NRT'' and ``RT'' indicate Non-Real-Time and Real-Time text detection methods, respectively. ``\textcolor{blue}{\textbf{Blue}}'' and ``\textcolor{red}{\textbf{Red}}'' are the best results of Non-Real-Time and Real-Time text detection methods, respectively. ``*'' means the result measured in our environment.}
	\label{msra}
	\vspace{-3mm}
\end{table}

\begin{figure*}
	\begin{center}
		\includegraphics[width=0.9\textwidth]{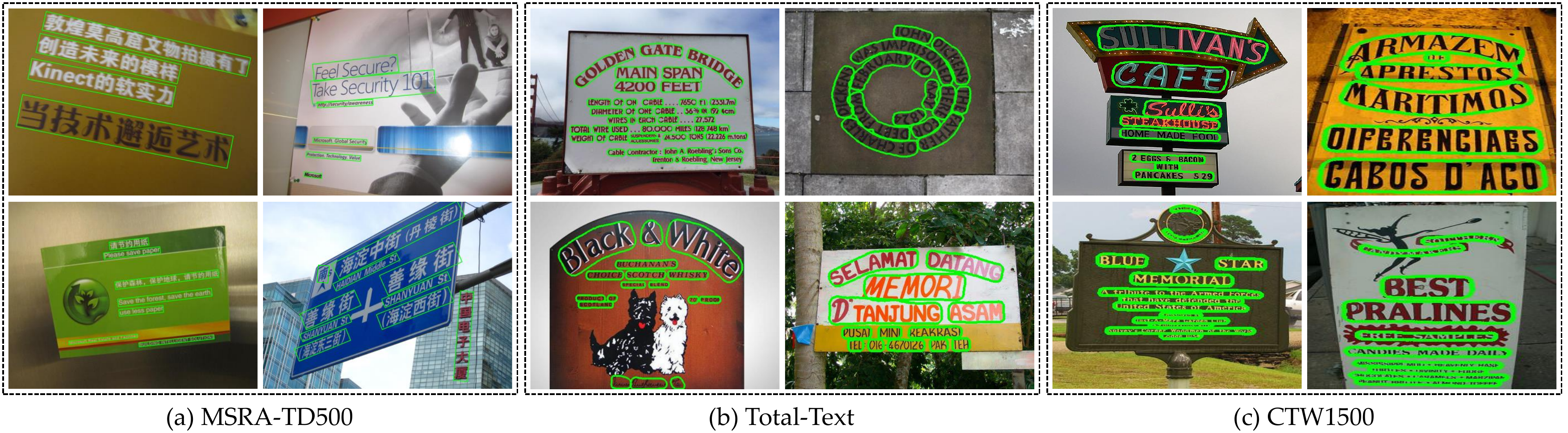}
	\end{center}
	\vspace{-6mm}
	\caption{Illustration of some qualitative detection results of the proposed ASMTD.}
	\label{V8}
\end{figure*}

\subsection{Comparison with State-of-the-Art Methods}
In this section, we compare our method with the related state-of-the-art (SOTA) methods on MSRA-TD500, Total-Text, and CTW1500 datasets to show the superiority of the ASMTD in both detection accuracy and speed.

\textbf{MSRA-TD500: Long Straight Text Benchmark.} The results on MSRA-TD500 are shown in Tab.~\ref{msra}. ASMTD achieves the F-measure of 86.3\% at an astonishing detection speed (62.5 FPS). For real-time text detection methods, our method outperforms DB~\cite{liao2020real} by 3.5\% about F-measure because text contours can be rebuilt accurately even when the predicted shrink-masks deviate from ground-truth. Moreover, the lightweight feature merging branch facilitates ASMTD to run 2x times faster than PAN~\cite{wang2019efficient}. Compared with OPMP~\cite{zhang2020opmp} and SAVTD~\cite{DBLP:conf/cvpr/FengYZL21}, since our method is equipped with a simpler framework and SPW, which helps to surpass them by 1.6\%--1.1\% in F-measure and runs 40x times faster than them at least. The detection results demonstrate the superiority of our method for detecting long straight text instances. We also illustrate some qualitative results in Fig.~\ref{V8}~(a).

\begin{table}
	\centering
	\setlength{\tabcolsep}{1.65mm}
	\begin{tabular}{c|c|cccc}
		\hline
		Type & Methods & P (\%) & R (\%) & F (\%)& FPS \\ \hline
		& EAST~\cite{zhou2017east} & 50.0 & 36.2 & 42.0 & - \\ \cline{2-6}
		& TextSnake~\cite{long2018textsnake} & 82.7 & 74.5 & 78.4 & - \\ \cline{2-6} 
		& TextField~\cite{xu2019textfield} & 81.2 & 79.9 & 80.6 & - \\ \cline{2-6} 
		& TextRay~\cite{wang2020textray} & 83.5 & 77.9 & 80.6 & - \\ \cline{2-6} 
		& OPMP~\cite{zhang2020opmp} & 85.2 & 80.3 & 82.7 & 3.7 \\ \cline{2-6} 
		\multirow{2}{*}{NRT}
		& LOMO~\cite{zhang2019look} & 87.6 & 79.3 & 83.3 & - \\ \cline{2-6} 
		& FCENet~\cite{zhu2021fourier} & 87.4 & 79.8 & 83.4 & - \\ \cline{2-6} 
		& CRAFT~\cite{baek2019character} & 87.6 & 79.9  & 83.6 & - \\  \cline{2-6} 
		& ABCNet~\cite{liu2020abcnet} & 87.9 & 81.3 & 84.5 & - \\  \cline{2-6} 
		& ReLaText~\cite{ma2021relatext} & 86.0 & 83.3 & 84.8 & \textcolor{blue}{\textbf{10.6}} \\ \cline{2-6} 
		& ContourNet~\cite{wang2020contournet} & 86.9 & 83.9 & 85.4 & 3.8 \\ \cline{2-6} 
		& DRRG~\cite{DBLP:conf/cvpr/ZhangZHLYWY20} & 86.5 & 84.9 & \textcolor{blue}{\textbf{85.7}} & - \\ \hline
		& DB~\cite{liao2020real}& 88.3 & 77.9 & 82.8 & 47.7*  \\ \cline{2-6} 
		RT
		& PAN~\cite{wang2019efficient} & 89.3 & 81.0 & 85.0 & 39.6 \\ \cline{2-6}
		& \textbf{Ours} & 88.5 & 83.8 & \textcolor{red}{\textbf{86.1}} & \textcolor{red}{\textbf{70.9}} \\ \hline
		
	\end{tabular}
	\vspace{-3mm}
	\caption{Comparison with related methods on Total-Text. ``NRT'' and ``RT'' indicate Non-Real-Time and Real-Time text detection methods, respectively. ``\textcolor{blue}{\textbf{Blue}}'' and ``\textcolor{red}{\textbf{Red}}'' are the best results of Non-Real-Time and Real-Time text detection methods, respectively. ``*'' means the result measured in our environment.}
	\label{tt}
	\vspace{-3mm}
\end{table}

\begin{table}
	\centering
	\setlength{\tabcolsep}{1.65mm}
	\begin{tabular}{c|c|cccc}
		\hline
		Type & Methods & P (\%) & R (\%) & F (\%)& FPS \\ \hline
		& EAST~\cite{zhou2017east} & 78.7 & 49.1 & 60.4 & \textcolor{blue}{\textbf{21.2}} \\ \cline{2-6}
		& TextSnake~\cite{long2018textsnake} & 67.9 & 85.3 & 75.6 & 1.1 \\ \cline{2-6} 
		& TextField~\cite{xu2019textfield} & 83.0 & 79.8 & 81.4 & - \\ \cline{2-6} 
		& TextRay~\cite{wang2020textray} & 82.8 & 80.4 & 81.6 & - \\ \cline{2-6} 
		& OPMP~\cite{zhang2020opmp} & 85.1 & 80.8 & 82.9 & 1.4 \\ \cline{2-6} 
		\multirow{2}{*}{NRT}
		& LOMO~\cite{zhang2019look} & 85.7 & 76.5 & 80.8 & - \\ \cline{2-6} 
		& FCENet~\cite{zhu2021fourier} & 85.7 & 80.7 & 83.1 & - \\ \cline{2-6} 
		& CRAFT~\cite{baek2019character} & 86.0 & 81.1 & 83.5 & - \\  \cline{2-6} 
		& ABCNet~\cite{liu2020abcnet} & 81.4 & 78.5 & 81.6 & - \\  \cline{2-6} 
		& ReLaText~\cite{ma2021relatext} & 84.8 & 83.1 & 84.0 & - \\ \cline{2-6} 
		& ContourNet~\cite{wang2020contournet} & 83.7 & 84.1 & 83.9 & 4.5 \\ \cline{2-6} 
		& DRRG~\cite{DBLP:conf/cvpr/ZhangZHLYWY20} & 85.9 & 83.0 & \textcolor{blue}{\textbf{84.4}} & - \\ \hline
		& DB~\cite{liao2020real}& 84.8 & 77.5 & 81.0 & 53.1*  \\ \cline{2-6} 
		RT
		& PAN~\cite{wang2019efficient} & 86.4 & 81.2 & 83.7 & 39.8 \\ \cline{2-6}
		& \textbf{Ours} & 87.8 & 80.3 & \textcolor{red}{\textbf{83.9}} & \textcolor{red}{\textbf{72.1}} \\ \hline
		
	\end{tabular}
	\vspace{-3mm}
	\caption{Comparison with related methods on CTW1500. ``NRT'' and ``RT'' indicate Non-Real-Time and Real-Time text detection methods, respectively. ``\textcolor{blue}{\textbf{Blue}}'' and ``\textcolor{red}{\textbf{Red}}'' are the best results of Non-Real-Time and Real-Time text detection methods, respectively. ``*'' means the result measured in our environment.}
	\label{ctw}
	\vspace{-1mm}
\end{table}

\textbf{Total-Text: Word-Level Curved Text Benchmark.} As we can see from Tab.~\ref{tt}, ASMTD achieves 86.1 in F-measure and 70.9 in FPS, which outperforms previous state-of-the-art (SOTA) methods. Specifically, since the SPW provides rich semantic context for prediction tasks, the ASMTD outperforms ReLaText~\cite{ma2021relatext} and DRRG~\cite{DBLP:conf/cvpr/ZhangZHLYWY20} by 2.7\% and 0.4\% in F-measure. At the same time, ASM based text representation method saves much computational cost for the inference process, which makes our method can run 6.7 times faster than the fastest non-real-time method (ReLaText~\cite{ma2021relatext}). The results demonstrate the effectiveness of ASMTD for detecting word-level curved text instances. Moreover, as shown in Fig.~\ref{V8}~(b), though many text instances are close to each other, our method still can distinguish them accurately because of the superiority of ASM based text representation method.

\textbf{CTW1500: Line-Level Curved Text Benchmark.} As shown in Tab.~\ref{ctw}, since ASM helps ASMTD to enjoy the same efficiency as existing real-time methods and the feature merging branch reduces the network complexity, our model runs 19 FPS and 32.3 FPS faster than DB and PAN, respectively. Moreover, SPW brings significant improvement for recognizing shrink-masks by providing rich context information and defining the semantic difference with interval region, which makes ASMTD outperform most non-real-time methods in F-measure. Although our method is a little lower (0.5\%) than DRRG~\cite{DBLP:conf/cvpr/ZhangZHLYWY20} in F-measure, ASMTD has a least 70x times faster speed (72.1 FPS) than it. The evaluation results on the CTW1500 show the model robustness for detecting long curved text instances. Moreover, some qualitative detection results on CTW1500 are illustrated. As we can see from Fig.~\ref{V8}~(c)~first row and second column, even there are some interference regions that enjoy similar low-level features (such as color and texture), our method still can distinguish them from texts effectively.

\begin{table}
	\centering
	\setlength{\tabcolsep}{3.8mm}
	\begin{tabular}{c|ccc}
		\hline
		Evaluation dataset                  & \multicolumn{3}{c}{Training on MSRA-TD500}                   \\ \hline
		\multirow{2}{*}{CTW1500}    & \multicolumn{1}{c}{P (\%)} & \multicolumn{1}{c}{R (\%)}    & F (\%)   \\ \cline{2-4} 
		& \multicolumn{1}{c}{82.7}  & \multicolumn{1}{c}{74.3} & 78.3 \\ \hline
		Evaluation dataset                 & \multicolumn{3}{c}{Training on CTW1500}                      \\ \hline
		\multirow{2}{*}{MSRA-TD500} & \multicolumn{1}{c}{P (\%)}     & \multicolumn{1}{c}{~~R (\%)}    & F (\%)    \\ \cline{2-4} 
		& \multicolumn{1}{c}{82.5}  & \multicolumn{1}{c}{77.8} & 80.1 \\ \hline
	\end{tabular}
	\vspace{-3mm}
	\caption{Cross-dataset evaluations on line-level datasets.}
	\label{cross}
\end{table}

\begin{table}
	\centering
	\setlength{\tabcolsep}{0.95mm}
	\begin{tabular}{c|c|ccc|cc}
		\hline
		\multirow{2}{*}{Dataset} & \multirow{2}{*}{S} & \multicolumn{3}{c|}{Time consumption (ms)} & \multirow{2}{*}{F(\%)} & \multirow{2}{*}{FPS} \\ \cline{3-5}
		&                       & Backbone        & Head        & Post       &                            &                      \\ \hline
		MSRA-TD500               & 736                   & 8.1            & 5.8         & 2.1        & 86.3                       & 62.5                 \\ 
		
		Total-Text               & 640                   & 7.1              & 5.1         & 1.9        & 86.1                       & 70.9                 \\ 
		CTW1500                  & 640                   & 7.0             & 5.0           & 1.9        & 83.9                       & 72.1                 \\ \hline
	\end{tabular}
	\vspace{-3mm}
	\caption{Time consumption of ASMTD on three public benchmarks. The total time
		consists of backbone, head and post-processing. ``S'' means that the shorter side of each testing image. ``Head'' contains the feature merging branch and prediction headers . ``Post'' represents post-processing.}
	\label{speed1}
\end{table}

\subsection{Cross Dataset Text Detection}
We further verify the generalization ability of the proposed ASMTD by cross evaluation experiments in this section. Since both MSRA-TD500 and CTW1500 are line-level benchmarks, we design two groups experiments on them. Specifically, we train our model on MSRA-TD500 and test it on CTW1500 at first. Then, the proposed ASMTD is trained on CTW1500 and evaluated on MSRA-TD500. As shown in Tab.~\ref{cross}, since the SPW provides rich semantic information for prediction tasks and ASM makes text contours can be reconstructed according to inaccurate shrink-masks, which facilitates our method outperforms the TextSnake~\cite{long2018textsnake} (in Tab.~\ref{ctw}) by 2.7\% in F-measure on CTW1500 and surpasses EAST~\cite{zhou2017east}, PixelLink~\cite{DBLP:conf/aaai/DengLLC18}, and TextSnake~\cite{long2018textsnake} (in Tab.~\ref{msra}) by 1.8\% F-measure at least on MSRA-TD500. The experiment results verify the proposed ASMTD enjoys strong generalization and robustness for diverse shaped text instances in different benchmarks.

\begin{table}
	\centering
	\setlength{\tabcolsep}{0.69mm}
	\begin{tabular}{c|c|ccc}
		\hline
		\multirow{2}{*}{Methods} & \multirow{2}{*}{GFLOPs} & \multicolumn{3}{c}{F-measure (\%)}                                              \\ \cline{3-5} 
		&                         & \multicolumn{1}{c|}{MSRA-TD500} & \multicolumn{1}{c|}{Total-Text} & CTW1500 \\ \hline
		PAN~\cite{wang2019efficient}                      & 63.88                  & \multicolumn{1}{c|}{84.1}           & \multicolumn{1}{c|}{85.0}           &     83.7    \\ 
		DB~\cite{liao2020real}                   & 52.54                   & \multicolumn{1}{c|}{82.8}           & \multicolumn{1}{c|}{82.8}           &   81.0      \\ 
		Ours                     &        46.96                & \multicolumn{1}{c|}{86.3}           & \multicolumn{1}{c|}{86.1}           &    83.9     \\ \hline
	\end{tabular}
	\vspace{-3mm}
	\caption{Comparison of computational cost and performance of different real-time detectors. ``GFLOPs'' indicates floating point of operations.}
	\label{speed2}
\end{table}

\begin{figure}
	\begin{center}
		\includegraphics[width=0.4\textwidth]{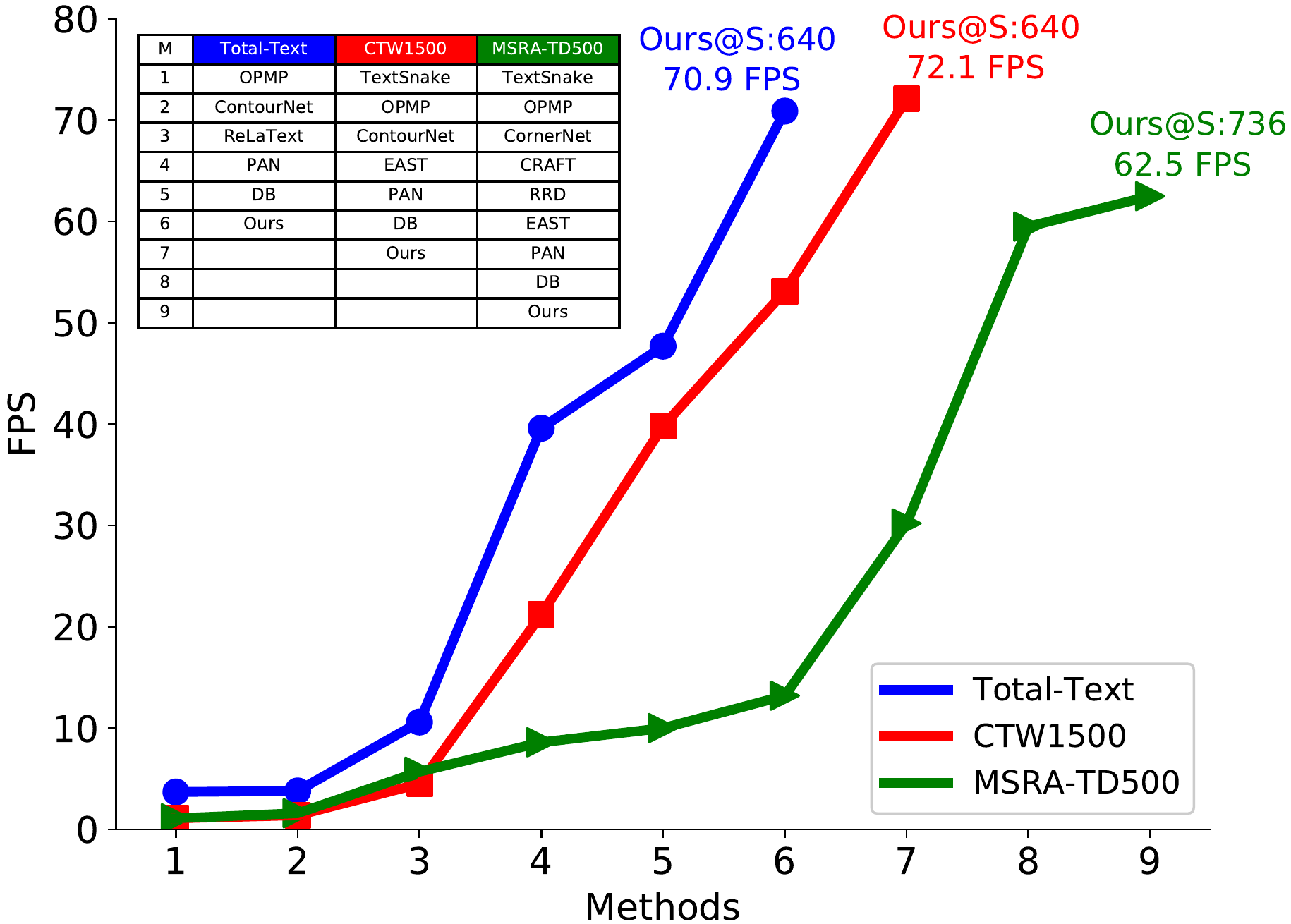}
	\end{center}
	\vspace{-5mm}
	\caption{Comparison of detection speed on multiple benchmarks. ``S'' is the shorter side of each testing image for our method.}
	\label{V9}
	\vspace{-3mm}
\end{figure}

\subsection{Speed Analysis}
To verify the superiority of our method in terms of detection speed and computational resources, we show the time consumption of ASMTD and compare the computational cost with related SOTA methods. As illustrated in Tab.~\ref{speed1}, since the ASM helps text contours can be rebuilt through extending shrink-mask contours by adaptive offsets directly, which makes the post-processing takes about 14\% of the total time consumption. At the same time, because the SPW can be removed from the inference stage and the lightweight feature merging branch saves much computational cost, the head only takes about 36\% of the total time consumption (as shown in Fig.~\ref{V3}). The experiment results demonstrate the effectiveness of the proposed ASM and SPW for improving the model detection speed. Additionally, we verify the superiority of ASMTD in detection speed. As shown in Fig.~\ref{V9}, our method outperforms other related SOTA methods a lot. Moreover, we show the computational cost of existing real-time methods in Tab.~\ref{speed2}. Because ASMTD only consists of two prediction headers and is equipped with a lightweight feature merging branch, our method enjoys the least floating point of operations (FLOPs) and the highest detection accuracy.

\section{Conclusion}
\label{Conclusion}
In this paper, we propose a novel real-time framework to detect arbitrary-shaped texts with high detection accuracy. We firstly propose the Adaptive Shrink-Mask (ASM) to fit text instances by shrink-masks and adaptive offsets, which improves the accuracy of rebuilt text contours effectively. Then, we introduce the Super-pixel Window (SPW) to provide rich semantic context for the prediction tasks, which brings significant improvements for the reliability of the discrimination of shrink-masks and the prediction of adaptive offsets. Importantly, SPW can be removed from the inference process and brings no extra computational cost. In the end, a lightweight feature merging branch is constructed to concatenate multi-level feature maps, which further facilitates the detection speed. Extensive experiments demonstrate the effectiveness of the AES and SPW. Comparison experiments show that the proposed ASMTD is superior to all state-of-the-art real-time methods.

{\small
\bibliographystyle{ieee_fullname}
\bibliography{egbib}
}

\end{document}